\documentclass[conference]{IEEEtran}
\usepackage{amsmath,amsfonts}
\usepackage{algorithmic}
\usepackage{algorithm}
\usepackage{array}
\usepackage[caption=false,font=normalsize,labelfont=sf,textfont=sf]{subfig}
\usepackage{textcomp}
\usepackage{stfloats}
\usepackage{url}
\usepackage{verbatim}
\usepackage{graphicx}
\usepackage{cite}
\usepackage[table,xcdraw]{xcolor}
\usepackage{soul}
\usepackage{tikz}
\usepackage{yquant}
\usepackage{blochsphere}
\usepackage{braket}
\usepackage{multirow}
\usepackage[mathscr]{euscript}
\usepackage{mathtools}
\usepackage{balance}
\DeclarePairedDelimiter{\floor}{\lfloor}{\rfloor}

\begin{document}

\title{A Novel Stochastic LSTM Model Inspired by Quantum Machine Learning}


\author{\IEEEauthorblockN{Joseph Lindsay, Ramtin Zand}
\IEEEauthorblockA{Department of Computer Science and Engineering, University of South Carolina, Columbia, SC 29208, USA\\
}
}

\markboth{Journal of \LaTeX\ Class Files,~Vol.~14, No.~8, August~2021}%
{Shell \MakeLowercase{\textit{et al.}}: A Sample Article Using IEEEtran.cls for IEEE Journals}


\maketitle

\bstctlcite{BSTcontrol}

\begin{abstract}
Works in quantum machine learning (QML) over the past few years indicate that QML algorithms can function just as well as their classical counterparts, and even outperform them in some cases. Among the corpus of recent work, many current QML models take advantage of variational quantum algorithm (VQA) circuits, given that their scale is typically small enough to be compatible with NISQ devices and the method of automatic differentiation for optimizing circuit parameters is familiar to machine learning (ML). While the results bear interesting promise for an era when quantum machines are more readily accessible, if one can achieve similar results through non-quantum methods then there may be a more near-term advantage available to practitioners. To this end, the nature of this work is to investigate the utilization of stochastic methods inspired by a variational quantum version of the long short-term memory (LSTM) model in an attempt to approach the reported successes in performance and rapid convergence. By analyzing the performance of classical, stochastic, and quantum methods, this work aims to elucidate if it is possible to achieve some of QML’s major reported benefits on classical machines by incorporating aspects of its stochasticity.
\end{abstract}

\vspace{-3mm}
\section{Introduction}\label{sec:intro}

An area of quantum computing research that has experienced popularity in recent years is the use of such systems for machine learning (ML) applications. Aside from the novelty of simply applying quantum computing to ML tasks, quantum machine learning (QML) overall has shown promise in providing a quantum advantage to several learning subroutines and models \cite{biamonte2017, dunjko2017, prati2017, tacchino2019, date2021, schuld2019-2, kerenidis2020, schuld2014, amin2018, anschuetz2019}. Although there exist various formulations to implement QML methods, many are being realized through the use of variational quantum algorithms (VQA). These VQAs function similarly to automatic differentiation techniques, such as what is commonly applied in neural network learning, with parametrized quantum operations being variationally updated and classically optimized such that the final measurement upon the circuit delivers the desired result \cite{mitarai2018, schuld2019, mari2021}. These algorithms enable a quantum advantage for several problems, including simulation of dynamical quantum systems, combinatorial optimization, and chemistry problems; however, the likeness to ML methods is what makes their application to such systems natural and exciting \cite{mcclean2016, cerezo2020}.

Even narrowing the scope to specifically variational QML, there has been a myriad of successful works in recent years to take as examples. A variational quantum Boltzmann machine was demonstrated by Zoufal et al. \cite{zoufal2021} that produces results and performance metrics competitive to all of the classic classifiers it was compared against. Junde et al \cite{junde2021} introduced a suite QML techniques to aid drug discovery that included a variational generative adversarial network. A recent work by Chen et al. \cite{chen2020} gives the construction and theory of a variational quantum long short-term memory (LSTM) model; the authors demonstrate through several experiments that it is capable of converging to optimums faster and generally achieving lower loss values than its classical version.

While the utilization of quantum computing methods appears to be of success in this manner, the advantage may not be the most accessible to practitioners and researchers. Cloud access to modern noisy intermediate-scale quantum (NISQ) devices \cite{bharti2021} can come with large communication overheads, and the use of quantum simulators likewise drives up model training and inference times over the theoretical levels. While there are many benefits that can be exclusive to the paradigm, it may be possible to incorporate a facet of its inherent stochasticity into otherwise classic models. On its own, stochasticity can provide many benefits to ML models, including greater robustness and the ability to generalize on sparse and non-discrete data, and be integrated via probabilistic devices. As such, a stochastic model with similar benefits demonstrated by its quantum counterpart may be a greater boon in the near-term. This brings to mind an important question: \textit{Can we achieve similar results to variational QML by incorporating stochasticity inspired by its architectures into classical machine learning models?} If this were the case, then perhaps some of these advantages could be achieved without the need for quantum hardware.


Herein, we select LSTM models as a case study through which we investigate the aforementioned research question. LSTM models are designed for analyzing long-term temporal dependencies and patterns in sequential data. This imparts a wide range of applications for the model, in areas such as natural language processing \cite{ghosh2016}, trend analysis \cite{pradhan2021}, speech recognition \cite{han2017}, anomaly detection \cite{malhotra2015}, and much more. With the variety of input/output configurations possible and the improvements upon standard recurrent neural network (RNN) architecture, the LSTM model has proven its merit in time-series analysis problems \cite{jozefowicz2015, olah2015, gers2000}. However, as Chen et al. \cite{chen2020} have shown, improvements over its conventional construction can be made by using emerging computation paradigms such as quantum and stochastic computing. In this work, we present a novel approach for a stochastic LSTM model inspired by the aforementioned quantum LSTM (QLSTM) and demonstrate its success in performance. 



The remainder of the paper is organized as follows. In Section \ref{sec:background}, we introduce some relevant background information on quantum computing, variational quantum algorithms, and the LSTM model for the general reader. Then a quantum version of the LSTM model is presented in Section \ref{sec:qlstm}, which is followed by our inspired approach for a stochastic LSTM model in Section \ref{sec:slstm}. Finally, experimental results and analyses of the models is presented in Section \ref{sec:results} before concluding the work with final discussions in Section \ref{sec:conclusion}.

\section{Background}\label{sec:background}
\subsection{Relevant Quantum Computing Basics for the Novice}\label{sec:qc}
Unlike reading the state of classical bits which may be done at any time with no consequence, the quantum mechanical nature of qubits prevents us from reading their precise state in the same manner. Regardless of the complex orientation of the qubits, a measurement operation will probabilistically collapse the system to one of the basis states for the measurement, in effect destroying its previous quantum state.

For example, if a single qubit were measured in the computational basis then its resulting state will become either $\ket{0}$ or $\ket{1}$. The probability of the system existing in any given basis state for the measurement can be determined by the complex amplitudes that comprise the superposition of its qubits. To be more concise, for a 1-qubit state we would have
\begin{equation}
	\ket{\psi} = 
	\begin{bmatrix}
		\alpha \\
		\beta
	\end{bmatrix}
	= \alpha
	\begin{bmatrix}
		1 \\
		0
	\end{bmatrix}
	+ \beta
	\begin{bmatrix}
		0 \\
		1
	\end{bmatrix}
	= \alpha \ket 0 + \beta \ket 1 \text{,}
\end{equation}
where $\alpha$ and $\beta$ are the complex amplitudes for the $\ket{0}$ and $\ket{1}$ computational basis states.

Without loss of generality, quantum states like $\ket{\psi}$ should be unit vectors in the complex vector space, where they are constrained by the normalization condition such that
\vspace{-1mm}
\begin{equation}
	|\alpha|^{2} + |\beta|^{2} = 1 \text{.}
    \vspace{-1mm}
\end{equation}
This can also be simplified and furthermore extended to multi-qubit systems, such that
\vspace{-1mm}
\begin{equation}
	\sum_i |\alpha_{i}|^{2} = 1 \text{.}
    \vspace{-1mm}
\end{equation}
Given this, the square magnitude of a state's complex amplitude can be interpreted as the probability of observing the quantum computer in the corresponding state. 

To handle this manner of stochastic measurement, an expectation value of repeated measurements will be taken as the result. While experimentally this could be described as an average of all possible outcomes weighted by their probabilities, we can describe a mathematical formulation to provide an analytical, or exact, expectation. Given a quantum system in the state $\ket{\psi}$ and some valid quantum operator $U$, we can define the expectation value of measuring $U$ as
\begin{equation}\label{eq:expval}
	\braket{U}_{\psi}:=\bra{\psi} U \ket{\psi}\text{.}
\end{equation}
Here we call $U$ an observable, with the possible results from performing a measurement on $U$ being its eigenvalues.

By applying spectral decomposition, we can view $U$ in terms of its eigenvalues $\lambda_i$ and eigenvectors $\ket{\phi_i}$ to elucidate how this produces the "expected result". To be concise, this formalism can be mathematically described as
\begin{equation}
	U = \sum_i \lambda_i \ket{\phi_i}\bra{\phi_i}\text{.}
\end{equation}

Rewriting Equation \ref{eq:expval} using this formulation of $U$,
\begin{equation}
	\begin{split}
		\braket{U}_{\psi} &=\bra{\psi} U \ket{\psi}\text{,} \\
		&=\sum_i \lambda_i \braket{\psi | \phi_i}\braket{\phi_i | \psi}\text{,} \\
		&=\sum_i \lambda_i \mid\braket{\phi_i | \psi}\mid^2\text{,}
	\end{split}
\end{equation}
where $|\braket{\phi_i | \psi}|^2$ gives the probability of a measurement resulting in $\lambda_i$ when the system is in state $\ket{\psi}$ \cite{nielsen2019, kaye2010, watrous2018}.

This computational scheme is commonly the default for measurement operations in quantum simulators, but conversely cannot be applied when measuring a physical quantum device since its qubits' precise orientations cannot be known without collapsing their states. Therefore, many simulators, such as IBM's Qiskit \cite{Qiskit} and Xanadu's PennyLane \cite{bergholm2020}, will allow a "shot" parameter to be specified so that the true stochastic behavior of measurement operations can be emulated.


\subsection{Variational Quantum Algorithms}\label{sec:vqa}
VQA systems exist as hybrid systems that can achieve a great majority of the envisioned uses for quantum computing via leveraging quantum circuits with tunable parameters \cite{cerezo2020}. Use of a technique known as the parameter-shift rule, briefly summarized as the difference in expectation between two circuits where a given parameter is shifted by a constant derived from unique eigenvalues, enables classical optimization techniques to tune the circuit parameters. While this allows us to analytically find the quantum circuit gradients, common optimization methods are sufficient for the classical optimization of circuit parameters provided that there is a valid cost function for the problem to be solved \cite{mitarai2018, crooks2019, schuld2019, mari2021}. As such, the circuits for these VQAs, referred to as variational quantum circuits (VQC) for ease, are composed of three primary components: an ansatz suited to solving the task that has been encoded, an optimization procedure over the tunable parameters' gradients, and a cost function that's effective for the problem.

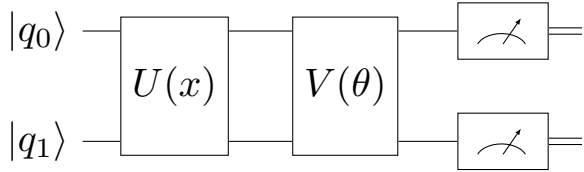
\begin{figure}
 	\centering
    \begin{tikzpicture}[scale=1.5]
        \begin{yquant}
            qubit {$\ket{\reg_{\idx}}$} q[2];
            box {$U(x)$} (q);
            box {$V(\theta)$} (q);
            measure q[-1];
        \end{yquant}
    \end{tikzpicture}
    \caption{A generalized example of a two-qubit VQC.}\label{fig:vqa_circuit}
\end{figure}

The most crucial portion of this is the ansatz, which consists of a sequence of quantum operations that are either proposed using information about the problem or formulated to be as general as possible. This comprises the body of the quantum circuit for the VQA, defining its structure and encoding the parameters to be trained. Such a construction allows the quantum circuit depth to be kept shallow, making it extremely suitable to NISQ-era devices. While they have shown promise even at shallow depths, it should be noted that the theory of the field suggests that as many unitary gates for data encoding and the primary variational ansatz as necessary can be applied. A very general example of what such a circuit for a VQA may look like is illustrated in Fig. \ref{fig:vqa_circuit}; here the data $x$ is first encoded by unitary gate $U$, then the variational ansatz is applied through unitary $V$ with the parameter $\theta$. After the measurement operation, the parameter will undergo classical optimization with the selected optimizer and cost function \cite{cerezo2020, chen2020, verdon2019}. Some applications for this technique include finding ground and excited states of molecules, classical optimization procedures, quantum program compilation, and even quantum error correction, among many others \cite{cerezo2020, verdon2019}. Given the method's formulation, its easy to see how it can be considered as an analog to classical ML; thus, it is natural to consider extending it to such problems in data science, which is the use case for the work at hand. Research in this area has shown that QML models which utilize VQAs are successful in a variety of problems, can have greater expressive power, converge to optimal solutions faster, and can even produce better results \cite{biamonte2017, chen2020, zoufal2021}. 

\subsection{Long Short-Term Memory Model}\label{sec:lstm}
As a variant of the RNN, the LSTM model is best suited for analyzing time-series data sequences to learn temporal dependencies. The primary differences that make LSTM networks distinct from standard RNNs are the set of equations used within the network's cell and the addition of another parameter referred to as the cell state, passed to subsequent time steps of the network's current batch like its hidden state. These changes effectively address issues that standard RNNs have with vanishing gradients and allow the model to learn longer sequential dependencies in the data. Where $v_{t}$ is the previous hidden state concatenated with the current data s.t. $v_{t}=[h_{t-1},x_{t}]$, $W_f$ and $b_f$ are the weights and biases for the forget operation, $W_i$ and $b_i$ are the weights and biases for the input operation, $W_{\tilde{C}}$ and $b_{\tilde{C}}$ are the weights and biases for the update candidate operation, and $W_o$ and $b_o$ are the weights and biases for the output operation, a classic LSTM cell can be mathematically expressed by the following equations \cite{hochreiter1997long}:

\vspace{-2mm}
\begin{subequations}\label{eq:lstm}
	\begin{equation}\label{eq:forget}
		f_{t}=\sigma(W_{f} \cdot v_{t} + b_{f})
	\end{equation}
	\begin{equation}\label{eq:input}
		i_{t}=\sigma(W_{i} \cdot v_{t} + b_{i})
	\end{equation}
	\begin{equation}\label{eq:update}
		\tilde{C}_{t}=tanh(W_{\tilde{C}} \cdot v_{t} + b_{\tilde{C}})
	\end{equation}
	\begin{equation}\label{eq:cell}
		c_{t}=(f_{t} * c_{t-1}) + (i_{t} * \tilde{C}_{t})
	\end{equation}
	\begin{equation}\label{eq:ouput}
		o_{t}=\sigma(W_{o} \cdot v_{t} + b_{o})
	\end{equation}
	\begin{equation}\label{eq:hidden}
		h_{t}=o_{t} * tanh(c_{t})
	\end{equation}
\end{subequations}

\begin{figure}
\centering
\includegraphics[width=3.4in]{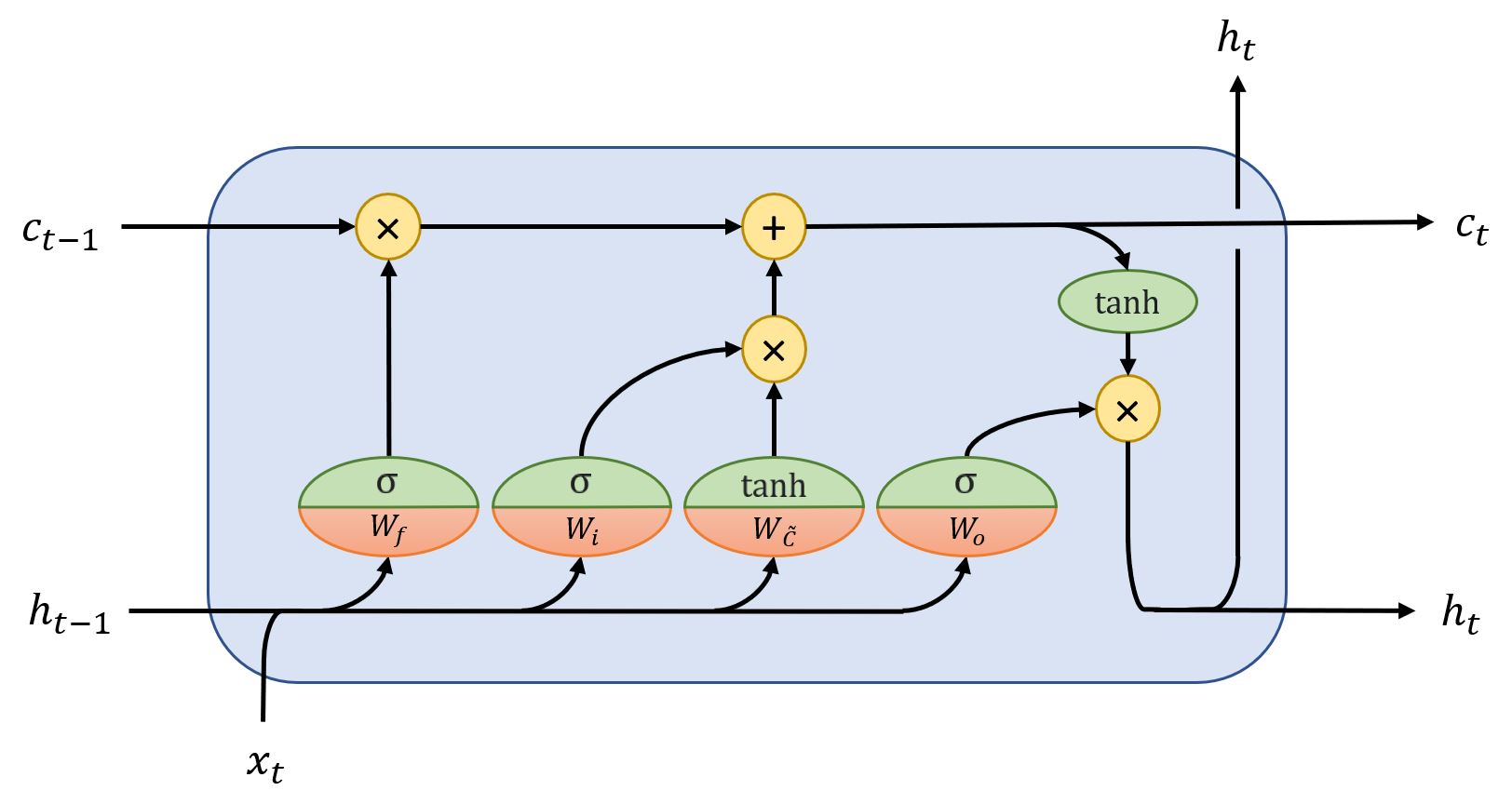}
\vspace{-3mm}
\caption{Graphical representation of an LSTM cell.}
\label{fig:lstm}
\end{figure}

This construction of an LSTM cell as described above is illustrated by Fig. \ref{fig:lstm}. It is worth mentioning that each of the multiply-and-accumulate (MAC) operations occur with separate sets of weights and biases, indicated by their subscripts. At the high level, the function of an LSTM cell can be described by the three gates it consists of: the forget gate, the input and update gate, and the output gate \cite{jozefowicz2015, olah2015, werbos1990, gers2000, chen2020}.
\begin{enumerate}
	\item Corresponding to Equation \ref{eq:forget} and the first part of Equation \ref{eq:cell}, the forget gate comes first and works to update the relevance of data from the previous cell state depending upon the current time step data. 
	\item Comprised by Equations \ref{eq:input}, \ref{eq:update}, and rest of \ref{eq:cell}, the input and update gate is the next portion of the LSTM cell and functions to decide what new information the current cell state should be updated with. 
	After determining which data values of the cell state should be updated via sigmoid activation and generating an initial candidate with which to update the cell state via hyperbolic tangent activation, the results are multiplied together element-wise to produce scaled candidate values to impart the update on the cell state resulting from the forget gate.
	\item Consisting of Equations \ref{eq:ouput} and \ref{eq:hidden}, the final portion of the LSTM cell, the output gate, sets the current hidden state of the network before passing it along with the cell state to the next time step and potentially returning that hidden state as output.
\end{enumerate}

\section{VQC-based Quantum LSTM}\label{sec:qlstm}
\begin{figure}
	\centering
	\includegraphics[width=3.4in]{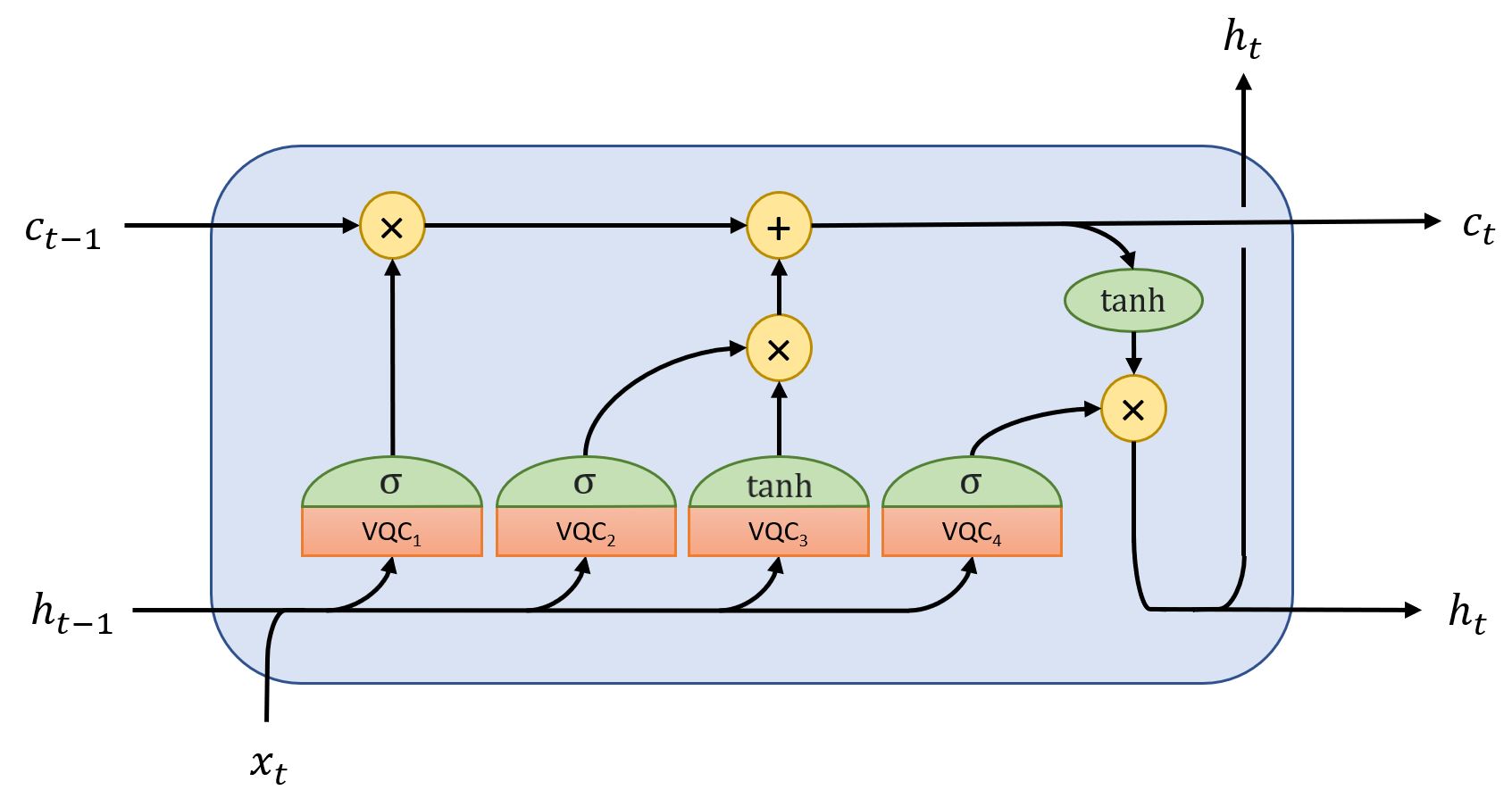}
 \vspace{-3mm}
	\caption{Graphical representation of a QLSTM cell.}
	\label{fig:qlstm}
\end{figure}

A quantum version of the LSTM architecture can be realized in a similar manner to its classical construction \cite{chen2020}. The primary difference is that the MAC operations within the LSTM cell have been replaced by VQCs. The authors of this work also applied the VQCs to the outgoing hidden state and LSTM cell output; however, an investigation done herein showed that this is not necessary for this QLSTM to achieve high performance. The QLSTM cell can then be illustrated by Fig. \ref{fig:qlstm}, and the model's equations become as follows:
\begin{subequations}
	\begin{equation}
		f_{t}=\sigma(VQC_{1} (v_{t}:\theta_f))
	\end{equation}
	\begin{equation}
		i_{t}=\sigma(VQC_{2} (v_{t}:\theta_i))
	\end{equation}
	\begin{equation}
		\tilde{C}_{t}=tanh(VQC_{3} (v_{t}:\theta_{\tilde{C}}))
	\end{equation}
	\begin{equation}
		c_{t}=(f_{t} * c_{t-1}) + (i_{t} * \tilde{C}_{t})
	\end{equation}
	\begin{equation}
		o_{t}=\sigma(VQC_{4} (v_{t}:\theta_o))
	\end{equation}
	\begin{equation}
		h_{t}=o_{t} * tanh(c_{t})
	\end{equation}
\end{subequations}

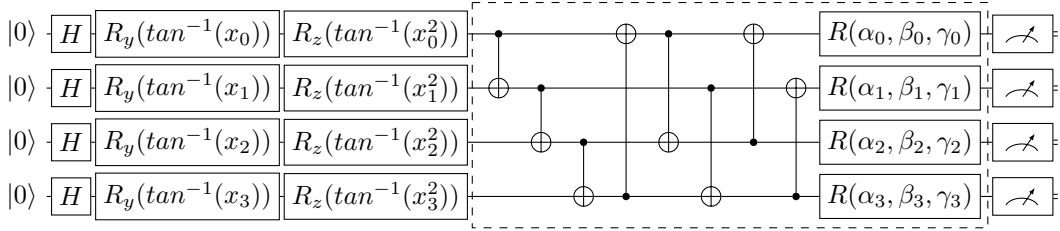
\begin{figure*}[!t]
 	\centering
    \begin{tikzpicture}
        \begin{yquant}[operator/separation=0.65mm]
            qubit {$\ket{0}$} q[4];
            h q[-3];
            box {$R_y(tan^{-1}(x_{\idx}))$} q[-3];
            box {$R_z(tan^{-1}(x_{\idx}^{2}))$} q[-3];
            [this subcircuit box style={dashed}]
            subcircuit {
                qubit {} q[4];
                cnot q[1] | q[0];
                cnot q[2] | q[1];
                cnot q[3] | q[2];
                cnot q[0] | q[3];
                cnot q[2] | q[0];
                cnot q[3] | q[1];
                cnot q[0] | q[2];
                cnot q[1] | q[3];
                box {$R(\alpha_\idx,\beta_\idx,\gamma_\idx)$} q[-3];
            } (q[-3]);
            measure q[-3];
        \end{yquant}
    \end{tikzpicture}
    \caption{The QLSTM VQC architecture introduced \cite{chen2020}; the variational ansatz in the dashed box can be applied as many times as necessary.}\label{fig:qlstm_vqc}
\end{figure*}

The VQC utilized for this construction has an architecture as shown by Fig. \ref{fig:qlstm_vqc}. This can be described as existing in three portions: a data encoding layer, a variational layer, and a measurement layer. The circuit starts by placing the initial ground state qubits into an unbiased state through a bank of Hadamard gates. The input data is then encoded as rotational angles of the qubits by transforming its values using the arctan function and applying the R$_{\text{y}}$ and R$_{\text{z}}$ gates.

The variational layer, indicated by the dashed box in the circuit figure, then generates multi-qubit entanglement through two rings of CNOT gates. Following this, the tunable parameters $\alpha$, $\beta$, and $\gamma$ of each qubit are encoded using single qubit rotational gates, i.e. R($\alpha$, $\beta$, $\gamma$). This variational layer may be applied multiple times, defining a depth hyperparameter to the circuit; for this work, only a single of such layers has been used. The final portion, as any quantum circuit must to return results, then measures each qubit. This is done on a computational basis as Pauli-Z expectation values.

\section{Stochastic LSTM}\label{sec:slstm}
\subsection{Stochasticity via Classical Quantization}\label{sec:quantization}
Weight quantization is a technique that has primarily been utilized for deploying ML models on low-precision devices that have limited computing resources and memory capacity, such as embedded systems and mobile devices \cite{QNN1,bnn1,bnn3,xnornet,mobile}. The data used by these systems are typically constrained to a smaller number of constituent bits than the real-valued high-precision data allowed by standard computing hardware. Furthermore, floating-point arithmetic may not even be possible on some of these devices. Quantization thus functions to logically reduce data bitwidth, as opposed to simply truncating the excess bits. These techniques generally convert floating-point representations to either fixed-point numbers or discrete integers; here, we only consider the case where resulting values are integers. Applications for this method can be broadly categorized into cases where a pre-trained model is quantized and cases where a model is quantized during its training, with the latter being the focus of this work. 

There are a few quantization methods that have proven effective, although naïve approaches and using a small number of resulting bits tend to produce poor results. This is largely due to small updates during optimization being rounded off, causing training to stagnate; this is similar to the vanishing gradient issue in training full-precision models, and can effectively have the same impact on optimization. The method used here is stochastic rounding quantization, which is both a common choice for weight quantization and suitable for the work at hand. While deterministic strategies are more prone to the aforementioned optimization issue, stochastic rounding is less affected by it given the method's probabilistic nature.

The mathematical formulation for this method is presented in Equation \ref{eq:sr}, where $Q_{s}(\cdot)$ is the stochastic rounding function, $w$ is the weight to be quantized, and $p$ is a uniformly generated random number \cite{li2017}. These functions are commonly followed by a clipping or clamping, function to ensure the result is within the range specified by the bit-width \cite{courbariaux2016}.

\begin{equation}\label{eq:sr}
	Q_{s}(w) =
	\begin{cases}
		\floor{w} + 1, & \text{for } p \leq w - \floor{w}\\
		\floor{w}, & \text{otherwise}
	\end{cases}
\end{equation}


\subsection{A Quantum-Inspired Stochastic Method}\label{sec:qmac}

The strategy for incorporating stochasticity into classical models utilized by this work is realized through the use of stochastic rounding quantization demonstrated by Equation \ref{eq:sr}. As presented in Section \ref{sec:quantization}, the stochastic benefits of this approach are limited to the training phase of a model. Despite this, quantization can be repurposed to incorporate stochasticity into the inference phase by generalizing its application beyond just model weights. Here, we propose to apply the stochastic rounding method to the output of MAC operations within ML model layers during the inference. This is inspired by the manner in which stochasticity is introduced into the quantum model described in Section \ref{sec:qlstm}; the measurement of the quantum circuit will probabilistically collapse its state after its main operations have occurred. Since the circuit is intended to replace the MAC operations of a standard LSTM model, emulating the paradigm's manner stochasticity into a classic model should be realized by augmenting the feed-forward path after the MAC operation and before the application of the corresponding activation function. This approach can be mathematically described as 
\begin{equation}\label{eq:qmac}
	\hat{u_{t}} = Q_s(u_{t}) = Q_s(W \cdot v_{t} + b) \text{ ,}
\end{equation}
where $u_{t}$ refers to the MAC output before the application of a quantization function, $\hat{u_{t}}$ is the quantized MAC output, and $W$ and $b$ are respectively the weights and biases. By utilizing this equation for all MAC operations within the LSTM cell, stochasticity can be instilled into the model's feed-forward path and, as such, within its inference phase.

This technique does not alter much of the construction when applied to a classic LSTM model. Rather than reconfiguring core aspects of the architecture like the QLSTM model presented in Section \ref{sec:qlstm}, this can be viewed more as a stochastic augmentation to the standard LSTM model. It should be noted that this alteration is only present in the feed-forward path of the model. By applying the ideas introduced and Equation \ref{eq:qmac}, the model equations then become:
\begin{subequations}\label{eq:slstm}
	\begin{equation}
		f_{t}=\sigma(Q_{s}(W_{f} \cdot v_{t} + b_{f}))
	\end{equation}
	\begin{equation}
		i_{t}=\sigma(Q_{s}(W_{i} \cdot v_{t} + b_{i}))
	\end{equation}
	\begin{equation}
		\tilde{C}_{t}=tanh(Q_{s}(W_{\tilde{C}} \cdot v_{t} + b_{\tilde{C}}))
	\end{equation}
	\begin{equation}
		c_{t}=(f_{t} * c_{t-1}) + (i_{t} * \tilde{C}_{t})
	\end{equation}
	\begin{equation}
		o_{t}=\sigma(Q_{s}(W_{o} \cdot v_{t} + b_{o}))
	\end{equation}
	\begin{equation}
		h_{t}=o_{t} * tanh(c_{t})
	\end{equation}
\end{subequations}

\section{Simulation Results}\label{sec:results}
\begin{figure*}[!t]
	\centering
	\subfloat[]{\includegraphics[width=1.55in]{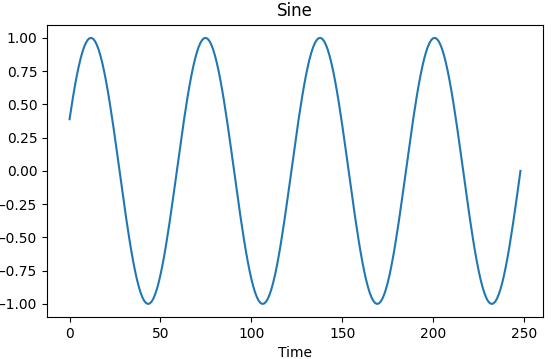}%
    \label{fig:sin}}
    \hfil
    \subfloat[]{\includegraphics[width=1.55in]{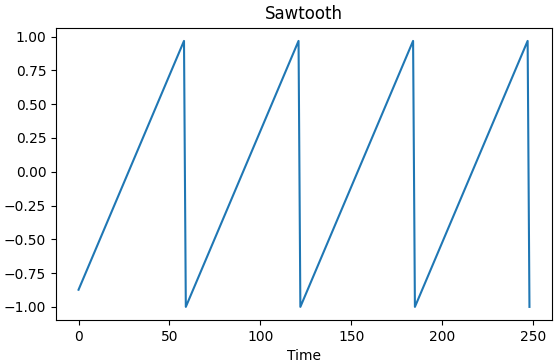}%
    \label{fig:saw}}
    \hfil
    \subfloat[]{\includegraphics[width=1.55in]{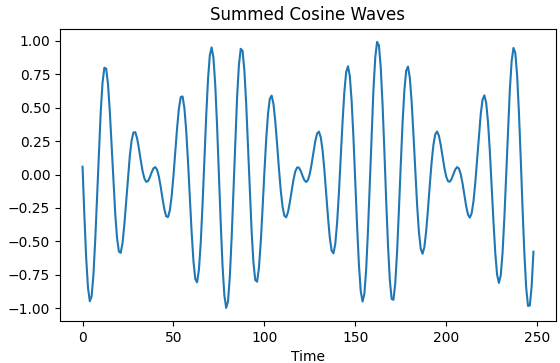}%
    \label{fig:wave_sum}}
    \hfil
	\subfloat[]{\includegraphics[width=1.55in]{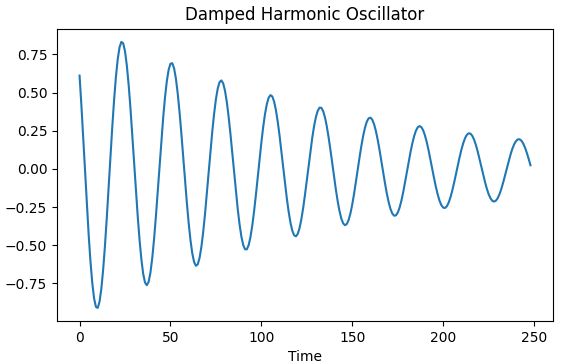}%
    \label{fig:osc}}
	\caption{Experiments performed herein to assess the models created: (a) \textit{sine}, (b) \textit{sawtooth}, (c) \textit{summed waves}, (d) \textit{damped harmonic oscillator} functions.}\label{fig:simulation_datasets}
\end{figure*}

\subsection{Experiments}\label{sec:exp}
Here, we have chosen four time-series data to benchmark the model performances. All selected signals can be broadly classified as either periodic or inspired by physical dynamics, and are kept simple to ease the workload for the quantum models. The first simulation dataset chosen was the trigonometric \textit{sine} function, shown in Fig. \ref{fig:sin}. Due to this pattern's simplicity, it it commonly used to confirm the validity of LSTM implementations and has been adopted herein as such.

While maintaining the need for general simplicity in pattern, the desire to introduce greater levels of intricacy over \textit{sine} led to the selection of the \textit{sawtooth} and \textit{summed cosine waves} periodic functions, with waveforms shown in Figures \ref{fig:saw} and \ref{fig:wave_sum} respectively. The former of these is a non-smooth and  non-continuous periodic function taken over four periods, with its linear nature disrupted between periods posing an interesting challenge. The latter of the aforementioned functions increases complexity by incorporating an additional layer of mathematics onto the standard trigonometric cosine; this formulation can be described by the equation:
\vspace{-1mm}
\begin{equation}\label{eq:wavepkt}
	y(x)=A_{1}cos(\frac{2 \pi x}{\lambda_{1}})+A_{2}cos(\frac{2 \pi x}{\lambda_{2}}) \text{,}
    \vspace{-1mm}
\end{equation}
where $A_1$ and $A_2$ refer to the two waves' amplitudes, while $\lambda_1$ and $\lambda_2$ are the waves' wavelengths, with the parameters used in the configuration for this work being $A_{1}=A_{2}=1$, $\lambda_{1}=9$, and $\lambda_{2}=11$.

Although this last one can be viewed through physics as the concept of beats, all functions introduced are periodic. To gauge model performance with a non-periodic function, the physical dynamics \textit{damped harmonic oscillator} system has been selected as the final simulation experiment. The motion of such a system can be defined by the differential equation
\vspace{-1mm}
\begin{equation}\label{eq:osc}
	\frac{d^{2}x}{dt^{2}}+2\chi\omega_{0}\frac{dx}{dt}+\omega_{0}^{2}x=0 \text{,}
    \vspace{-1mm}
\end{equation}
where $\chi$ is the damping ratio and $\omega_{0}$ is the characteristic frequency. This describes the physics that comprise a range of real systems, with the common example being a spring-based system that includes friction. In this case, we find that $\chi = \frac{c}{2\sqrt{mk}}$ and $\omega_{0} = \sqrt{\frac{k}{m}}$, where $m$ is the mass of the moving body, $k$ is the spring constant, and $c$ is the coefficient of friction. For this study, the values of $0.75$, $4$, and $0.1$ were chosen for $m$, $k$, and $c$, respectively. Fig. \ref{fig:osc} illustrates how the waveform for this may look.

These four experiments were used to benchmark and gather results from five different LSTM models. These model variants encompass: a standard, classic LSTM constructed in-house with Python 3.8.10; a QLSTM as described in Section \ref{sec:qlstm} constructed out-of-the-box using PennyLane \cite{bergholm2020} and PyTorch \cite{paszke2019}, utilizing simulator exact expectations; the same QLSTM, using simulator 1-shot expectation values to enable full stochastic results; a stochastic LSTM as described in Section \ref{sec:slstm} built on top of the in-house model, applying the quantization function a single time for 1-shot expectations; and the same stochastic LSTM, applying the quantization function 100 times before averaging to achieve a 100-shot expectation. 

All models were tested using the same hyperparameter settings. Omitting optimization level hyperparameters, the LSTM model settings used were an input dimension of $1$, a hidden dimension of $5$, an output dimension (for a final fully-connected layer after the LSTM cell) of $1$, and a batch size of $4$. Both reported quantum models utilized $4$ qubits and a variational depth of $1$, as shown in Fig. \ref{fig:qlstm_vqc}. The stochastic models used standard integer-level quantization for resolution. Additionally, it should be noted that all results presented were achieved using the RMSProp \cite{hinton2012} optimization procedure.

\subsection{Performance}\label{sec:performance}
Herein we present the performance of the five LSTM model variants across the four simulation experiments previously introduced. The training and validation RMSE values and R$^2$ scores from the training epochs with the lowest valued validation RMSE are given in Tables \ref{tab:best_sine}-\ref{tab:best_osc}. 

\begin{table}[hbt]
	\centering
	\caption{Best performance for all models on \textit{sine} experiments.}\label{tab:best_sine}
	\begin{tabular}{c||c|c||c|c}
		\hline
		\multicolumn{5}{c}{Sine} \\ \hline
		\multirow{2}{*}{Model} & \multicolumn{2}{c||}{Training} & \multicolumn{2}{c}{Validation}  \\
		\cline{2-5} & RMSE & R$^2$ & RMSE & R$^2$ \\
		\hline\hline
		In-house Classic & 0.0315 & 0.9969 & 0.0500 & 0.9933 \\ \hline
		Analytical Quantum & 0.0408 & 0.9944 & 0.0554 & 0.9914 \\ \hline
		1-Shot Quantum & 0.1729 & 0.8965 & 0.1718 & 0.8947 \\ \hline
		1-Shot Stochastic & 0.0717 & 0.9830 & 0.0715 & 0.9839 \\ \hline
		100-Shot Stochastic & 0.0430 & 0.9937 & 0.0522 & 0.9913 \\ \hline
	\end{tabular}
\end{table}

\begin{table}[hbt]
	\centering
	\caption{Best performance for all models on \textit{sawtooth} experiments.}\label{tab:best_sawtooth}
	\begin{tabular}{c||c|c||c|c}
		\hline
		\multicolumn{5}{c}{Sawtooth} \\ \hline
		\multirow{2}{*}{Model} & \multicolumn{2}{c||}{Training} & \multicolumn{2}{c}{Validation}  \\
		\cline{2-5} & RMSE & R$^2$ & RMSE & R$^2$ \\
		\hline\hline
		In-house Classic & 0.0484 & 0.8593 & 0.0905 & 0.7870 \\ \hline
		Analytical Quantum & 0.0465 & 0.8624 & 0.0907 & 0.7989 \\ \hline
		1-Shot Quantum & 0.2350 & 0.6124 & 0.2603 & 0.6041 \\ \hline
		1-Shot Stochastic & 0.0819 & 0.8450 & 0.1029 & 0.8200 \\ \hline
		100-Shot Stochastic & 0.0500 & 0.8618 & 0.0846 & 0.7833 \\ \hline
	\end{tabular}
\end{table}

\begin{table}[hbt]
	\centering
	\caption{Best performance for all models on \textit{summed cosine wave}.}\label{tab:best_wavesum}
	\begin{tabular}{c||c|c||c|c}
		\hline
		\multicolumn{5}{c}{Summed Waves} \\ \hline
		\multirow{2}{*}{Model} & \multicolumn{2}{c||}{Training} & \multicolumn{2}{c}{Validation}  \\
		\cline{2-5} & RMSE & R$^2$ & RMSE & R$^2$ \\
		\hline\hline
		In-house Classic & 0.0564 & 0.9797 & 0.0625 & 0.9761 \\ \hline
		Analytical Quantum & 0.0318 & 0.9917 & 0.0559 & 0.9817 \\ \hline
		1-Shot Quantum & 0.2217 & 0.7067 & 0.1973 & 0.7425 \\ \hline
		1-Shot Stochastic & 0.1184 & 0.9119 & 0.0897 & 0.9455 \\ \hline
		100-Shot Stochastic & 0.0720 & 0.9669 & 0.0735 & 0.9644 \\ \hline
	\end{tabular}
\end{table}

\begin{table}[hbt]
	\centering
	\caption{Best performance for all models on \textit{damped harmonic oscillator}.}\label{tab:best_osc}
	\begin{tabular}{c||c|c||c|c}
		\hline
		\multicolumn{5}{c}{Damped Harmonic Oscillator} \\ \hline
		\multirow{2}{*}{Model} & \multicolumn{2}{c||}{Training} & \multicolumn{2}{c}{Validation}  \\
		\cline{2-5} & RMSE & R$^2$ & RMSE & R$^2$ \\
		\hline\hline
		In-house Classic & 0.0140 & 0.9978 & 0.0199 & 0.9842 \\ \hline
		Analytical Quantum & 0.0120 & 0.9959 & 0.0049 & 0.9990 \\ \hline
		1-Shot Quantum & 0.1463 & 0.8186 & 0.0873 & 0.6081 \\ \hline
		1-Shot Stochastic & 0.1248 & 0.8665 & 0.0642 & 0.7810 \\ \hline
		100-Shot Stochastic & 0.0382 & 0.9861 & 0.0342 & 0.9414 \\ \hline
	\end{tabular}
\end{table}

The first thing to notice is that the in-house classical and analytical quantum models perform nearly on par for the experiments tested. The former of these produces slightly better results for \textit{sine}, while outclassed by the latter on physical dynamics-based experiments. Regardless, both showed high levels of success for every test they were applied to. This should be logical; these models deterministically follow a specific set of rules and equations. 

The story is different when the stochastic behavior of the quantum model is fully enabled. By evaluating its circuits with a single measurement sample rather than analytically computing the exact expectation, this model's performance significantly drops. The best validation RMSE values it manages to achieve is an order of magnitude higher for all trials than what is attained through use of exact expectations, with an average increase in validation RMSE of $+0.1275$. While the results are still acceptable, this configuration is outperformed by the in-house classical model in all experiments analyzed. 

By contrast, the degradation in performance for the 1-shot stochastic model compared to the classic model is relatively more modest; here we only see an average increase in the best validation RMSE by $+0.0042$. 
Although this model did not outperform either the classical or analytical quantum models, it demonstrated superior performance compared to the 1-shot quantum model across every experiment. The closest the performance for these two variants reached was a margin of $\pm0.0231$ for validation RMSE and $\pm0.1729$ for validation R$^2$ in the \textit{damped harmonic oscillator} function. Clearly, with full stochastic functionalities enabled, the quantum model cannot quite compete despite obtaining acceptable performance. 

\begin{figure}
	\centering
	\includegraphics[width=3.1in]{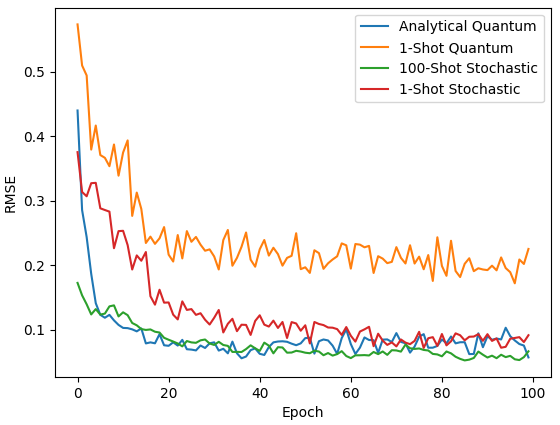}
	\vspace{-3mm}
	\caption{Plot of validation RMSE across training epochs for model variants to illustrate convergence on the \textit{sine} function.}
	\label{fig:sine_conv}
\end{figure}

Interestingly enough, the success demonstrated by the stochastic model is yet further improved by the utilization of the 100-shot expectation formulation. The performance improves in all experiments, a result of the stochastically quantized values converging to more optimal states similar to how we expect experimental quantum measurements with multi-shot expectation values to converge. Furthermore, the 100-shot stochastic model achieves the best validation RMSE in the \textit{sawtooth} simulations, attaining a value of $0.0846$ compared to the values of $0.0905$ and $0.0907$ that the classic and analytical quantum models, respectively, managed to reach. Although this was the experiment all models were least performant on, likely to due the asymptotic and abrupt change in linearity it contains, this provides insight into the improvement this method can offer to performance. Even if this result is the consequence of a stochastic variation, this demonstrates the promise that the technique can impart when data is more sparse and less discrete than the simulations done in this work.

Furthermore, the stochastic models demonstrate a capability to converge at a rate comparable to that demonstrated by the quantum models. Illustrated in Fig. \ref{fig:sine_conv} via a plot of validation RMSE values over the course of model training for \textit{sine}, both the quantum and stochastic models rapidly converge to their respective optimums in a highly similar manner. In this case, the stochastic models even outpaced the quantum models. More importantly, in either case, the models' convergence rates outpace that of the classical model. Although all trials used $100$ training epochs as a standard, this means that realistic applications for the model version would need to learn over fewer iterations to approach its best possible performances.

\subsection{Runtimes}\label{sec:runtimes}

It is worth briefly discussing the total runtime for the programs performing model training. 
Many promising works related to quantum results include theoretical asymptotic runtimes, but near-term results include a factor resulting from either quantum simulation or cloud computing overheads. The runtime taken by each model to produce the results analyzed, using an Intel® Core™ i7-6700HQ CPU @ 2.60GHz with 15.5 GB of memory, has been collected into Table \ref{tab:runtimes}. To reiterate, all of these times resulted from $100$ training epochs each using a purely classical machine. 

The shortest times are exhibited by the in-house and 1-shot stochastic models. While both are on the order of a single minute, the pseudo-random actions of the stochastic model slightly add to its time as should be expected. The 100-shot stochastic model compounds this additional runtime incurred by pseudo-random number generation, increasing to the order of about half an hour. Conversely to the rapid runtimes achieved by the classical models, the quantum realizations had much higher times to completion, which is expected considering that quantum dynamics are hard to classically simulate. The 1-shot expectation trials, however, conflate this issue with that of pseudo-random number generation to achieve its stochastic results, taking the longest of any implementations to complete the training procedure.
\begin{table}[]
	\centering
	\caption{Runtime for all models across every experiment in seconds (s), minutes (m), or hours (h). }\label{tab:runtimes}
	\begin{tabular}{c|c|c|c|c}
		\hline
		\multicolumn{5}{c}{Model Runtimes} \\ \hline
		Model & Sine & Sawtooth & \begin{tabular}[c]{@{}c@{}}Summed\\ Waves\end{tabular} & Oscillator \\
		\hline\hline
		In-house Classic & 24.88s & 25.57s & 24.57s & 23.91s \\ \hline
		Analytical Quantum & 56.34m & 57.13m & 55.56m & 52.82m \\ \hline
		1-Shot Quantum & 4.816h & 4.563h & 4.530h & 4.356h \\ \hline
		1-Shot Stochastic & 48.38s & 49.66s & 49.52s & 48.82s \\ \hline
		100-Shot Stochastic & 30.97m & 28.21m & 30.52m & 28.42m  \\ \hline
	\end{tabular}
\end{table}

Given the increased runtime for stochastic models due to the random number generation actions, it is worth contemplating using intrinsically stochastic devices to realize the proposed stochastic LSTM models. For instance, the stochastic techniques of this work may be realized directly in hardware via magnetoresistive random-access memory (MRAM) units \cite{MRAMBSN1,MRAMBSN2,MRAMBSN3,pyle2019subthreshold}. The magnetic tunneling junctions (MTJ) of these devices provide an energy barrier between their high and low resistance states such that it may be used as nonvolatile memory to store information \cite{fundamentals}. If this energy barrier within the MTJ is reduced, however, thermal noise and fluctuations can cause the MRAM to stochastically switch between its states. Electrical circuit constructions using only three transistors and one such MTJ allow for a sigmoidal distribution of output voltages that depend upon the input voltage; this culminates in a massive energy and area advantage for implementing random number generators over what is possible with conventional circuits \cite{pbit1}. By pairing this physical realization with experimental expectation values as described by this work, a high-performance stochastic model can be achieved. 

\section{Conclusion}\label{sec:conclusion}

Through an empirical case study, i.e., the LSTM model, this work approached matters of classical ML and QML to examine the effect of stochasticity on performance. Implementation of in-house models enabled an analysis of the effects that stochastic techniques can produce. Further, the construction of a quantum model through the PennyLane library allowed a first-hand comparison of the potential that the paradigm can afford to the practice as a whole. We demonstrated that the proposed stochastic LSTM model can achieve performance comparable to that of a quantum version of the same model. With this, we once more consider the initial question of this investigation as it reaches its conclusion. 
Through the results provided in this paper, it is shown that \textit{it is possible to realize similar performance to variational QML models by applications of stochasticity to classical models.}

Despite these findings, an analysis of LSTM models may not inherently transfer its results to all existing neural network architectures. The hypothesis that variational QML's advantage is not a result of its stochastic nature should be verified on other quantum models as a matter of further work. There may indeed be some cases where fully stochastic measurement operations lead to better performance than relying upon their analytical solutions. In addition, methods of incorporating stochasticity into ML models discussed by this study can be expanded and further evaluated. 

\balance
\bibliographystyle{IEEEtran}
\bibliography{references}

\end{document}